\documentclass{article}

\PassOptionsToPackage{numbers, compress}{natbib}

\usepackage[preprint]{neurips_2024}

\usepackage[utf8]{inputenc} 
\usepackage[T1]{fontenc}    
\usepackage{hyperref}       
\usepackage{url}            
\usepackage{booktabs}       
\usepackage{amsfonts}       
\usepackage{nicefrac}       
\usepackage{microtype}      
\usepackage{xcolor}         
\usepackage{graphicx}
\usepackage{amsmath}
\usepackage{subfigure}

\title{FlexiDreamer: Single Image-to-3D Generation with FlexiCubes}

\author{%
  Ruowen Zhao$^{1,4}$\quad Zhengyi Wang$^{2,4}$\quad Yikai Wang$^2$\quad Zihan Zhou$^3$\quad Jun Zhu$^{2,4}$\\
 $^1$ University of Chinese Academy of Sciences \and
$^2$ Tsinghua University \quad $^3$ Xidian University \quad $^4$ ShengShu \\
\url{https://flexidreamer.github.io/}
}

\begin{document}

\maketitle

\begin{figure}[htb]
  \centering
  \vspace{-0.5cm}
  \includegraphics[width=1.0\textwidth]{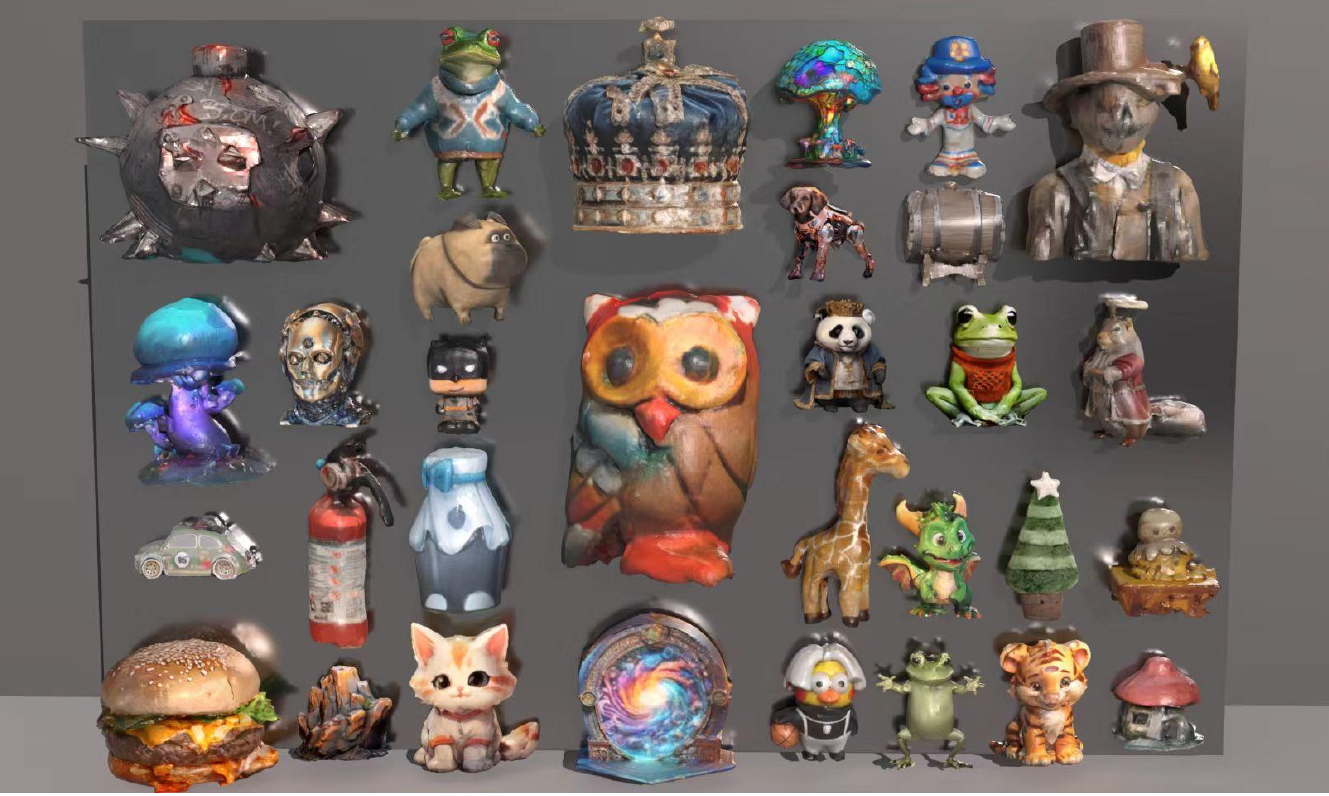}
  \centering
  \caption{
   \textbf{FlexiDreamer for single image-to-3D generation:} FlexiDreamer can reconstruct 3D content with detailed geometry and accurate appearance from a single image. We are able to generate a premium textured mesh in approximately one minute.
  }
  \label{fig:first}
\end{figure}

\begin{abstract}

  3D content generation has wide applications in various fields. One of its dominant paradigms is by sparse-view reconstruction using multi-view images generated by diffusion models. However, since directly reconstructing triangle meshes from multi-view images is challenging, most methodologies opt to an implicit representation (such as NeRF) during the sparse-view reconstruction and acquire the target mesh by a post-processing extraction. However, the implicit representation takes extensive time to train and the post-extraction also leads to undesirable visual artifacts. In this paper, we propose FlexiDreamer, a novel framework that directly reconstructs high-quality meshes from multi-view generated images. We utilize an advanced gradient-based mesh optimization, namely FlexiCubes, for multi-view mesh reconstruction, which enables us to generate 3D meshes in an end-to-end manner. To address the reconstruction artifacts owing to the inconsistencies from generated images, we design a hybrid positional encoding scheme to improve the reconstruction geometry and an orientation-aware texture mapping to mitigate surface ghosting. To further enhance the results, we respectively incorporate eikonal and smooth regularizations to reduce geometric holes and surface noise. Our approach can generate high-fidelity 3D meshes in the single image-to-3D downstream task with approximately 1 minute, significantly outperforming previous methods.

\end{abstract}

\section{Introduction}

3D content technologies allow us to create realistic digital representations of objects, environments and characters, enabling immersive experiences in virtual reality, gaming, animation and various other fields. Some approaches such as \cite{poole2022dreamfusion, chen2023fantasia3d, lin2023magic3d} distill 3D geometry with priors from powerful 2D diffusion model via Score Distillation Sampling (SDS), setting a new standard for integrating generalizability into 3D generation. However, these optimization-based methods often struggle with hours-long optimization time and over-saturated generation results.

Recently, multi-view diffusion has garnered significant attention due to its superior generalizability, quality, and efficiency, which facilitates the emergence of multi-view methods. By fine-tuning original 2D diffusion models on 3D dataset \cite{deitke2023objaverse,deitke2024objaverse}, multi-view methods first generate multi-view images of a 3D object based on an image or text prompt and then extract 3D geometry through multi-view reconstruction. Compared to SDS-based approaches, such methods can generate more consistent 3D geometric shapes without Janus problem and take only a few minutes to complete, significantly improving the efficiency of generation.

However, current multi-view diffusion models only generate sparse and inconsistent images owing to the limited computational resources. The sparse views and inconsistencies in the generated images add difficulty to reconstructing high-quality meshes. Since directly optimizing on mesh representation is challenging due to its difficult deformation operations, previous 3D generation works \cite{long2023wonder3d,liu2023syncdreamer,liu2023one2345,hong2023lrm,li2023instant3d} mostly train a NeRF-based representation, such as \cite{wang2023neus,long2022sparseneus} to reconstruct meshes from multi-view images. However, the training process of NeRF-based representation requires extensive time and resources, especially for high-resolution outputs. Additionally, it is unable to acquire the textured mesh in an end-to-end manner and relies on extra procedures such as Marching Cubes \cite{10.1145/37401.37422} to extract the iso-surface, often resulting in meshes with grid aliasing issues.


To overcome the limitations above, our key insight is to design an end-to-end mesh reconstruction framework from multi-view images. Therefore, we propose a novel mesh reconstruction framework, coined FlexiDreamer, by using an advanced gradient-based mesh optimization method FlexiCubes. With FlexiCubes, our approach can extract an explicit mesh from a learned signed distance function for iterative rendering and optimization, able to acquire textured mesh as the final output at the end of training. Furthermore, we render the mesh by surface rendering \cite{Laine2020diffrast}, which circumvents the slow rendering procedure in NeRF-based methods and diminishes generation time.

However, directly incorporating FlexiCubes into our reconstruction framework for 3D generation task is non-trivial. Firstly, given that there exist inconsistent details in multi-view images, we incorporate a hybrid positional encoding into the learning scheme of signed distance function to improve the geometry of reconstruction results. Then, due to the inaccuracies in the overlap area of two adjacent images, we design an orientation-aware texture mapping strategy for the mesh to mitigate surface ghosting. Finally, since regularizations in FlexiCubes are limited for reconstructing high-quality results, we additionally employ  eikonal regularization and two smooth regularizations to avoid geometric holes and minimize noise of mesh surface.

In summary, FlexiDreamer is an end-to-end framework based on FlexiCubes that obtains the target 3D meshes without extra procedures. Combined with image-conditioned multi-view diffusion models, FlexiDreamer enables a rapid generation of photorealistic textured meshes from single-view images in approximately 1 minute  on a single NVIDIA A100 GPU.

\section{Related Works}

\subsection{3D Generation with Diffusion Models}
Recently, 2D diffusion models have achieved notable success in text-to-image generation \cite{rombach2022high}. However, extending it to 3D generation poses a significant challenge. Existing 3D native diffusion models such as \cite{jun2023shap,nichol2022pointe,gao2022get3d,gupta20233dgen,liu2023meshdiffusion,chou2023diffusionsdf,cao2023largevocabulary,chen2023singlestage,müller2023diffrf,wang2022rodin,zhao2023michelangelo,yariv2023mosaicsdf} can generate consistent 3D assets within seconds, but they struggle to achieve open-vocabulary 3D generation due to the limited availability of extensive 3D training datasets. On the other hand, inspired by the remarkable progress of text-to-image synthesis in 2D diffusion models, DreamFusion \cite{poole2022dreamfusion} first proposes to distill 3D geometry and appearance from 2D diffusion priors via Score Distillation Sampling (SDS). Later methods \cite{wang2023prolificdreamer,lin2023magic3d,chen2023fantasia3d,li2023sweetdreamer,tang2023makeit3d,chen2023it3d,lorraine2023att3d,raj2023dreambooth3d,wang2023animatabledreamer,seo2024let,melaskyriazi2023realfusion,wang2022score,sun2023dreamcraft3d,ding2023textto3d,EnVision2023luciddreamer} build on DreamFusion and further enhance the quality of generated outputs. To tackle potential issues such as the Janus problem, \cite{shi2023mvdream,wang2023imagedream,qiu2023richdreamer,ye2024dreamreward} strengthen the semantics of different views when generating multi-view images. Recently, LRM \cite{hong2023lrm, tochilkin2024triposr} proposes a transformer-based reconstruction model to predict NeRF representation from single image in 5 seconds. Following works \cite{wang2023pflrm,li2023instant3d,xu2023dmv3d} combine LRM with pose prediction or multi-view diffusion model to perform a rapid and diverse generation. Moreover, some approaches, such as \cite{chen2023gsgen,tang2023dreamgaussian,yi2023gaussiandreamer,zou2023triplane,yin20244dgen,ling2024align,chen2023gaussianeditor,ren2023dreamgaussian4d,zhang2023repaint123,xu2024grm,tang2024lgm,szymanowicz2023splatter,xu2024agg,mercier2024hexagen3d,tang2024lgm,wang2024crm}, choose Gaussian Splatting \cite{kerbl20233d} or FlexiCubes \cite{shen2023flexicubes} as an alternative 3D representation in reconstruction to avoid the costly volumetric rendering in NeRF.

\subsection{Sparse-view Reconstruction with Multi-view Generation}
The evolution of 2D lifting methods has facilitated the development of consistent multi-view image generation by diffusion models \cite{liu2023zero,shi2023zero123,yang2023consistnet,ye2023consistent1to3,lin2024consistent123,shi2023tosshighquality}. With multi-view images obtained, the target 3D asset can be recovered quickly by a sparse-view reconstruction. Zero-1-to-3 \cite{liu2023zero} incorporates camera pose transformations into 2D diffusion models and realizes image-conditioned novel view synthesis. One-2-3-45 \cite{liu2023one2345} trains a reconstruction framework in conjunction with Zero-1-to-3 and SparseNeuS \cite{long2022sparseneus}, realizing a rapid 3D generation. SyncDreamer \cite{liu2023syncdreamer} uses spatial volume and depth-wise attention in a 2D diffusion model to generate multi-view consistent images. Wonder3D \cite{long2023wonder3d} and Zero123++ \cite{shi2023zero123} extend to generate multi-view RGB and normal images with cross-domain attention layers integrated into diffusion models. These high-fidelity generated images can yield 3D reconstruction via NeuS \cite{wang2023neus} or other NeRF variants \cite{chen2022tensorf,shi2023zerorf,fridovichkeil2023kplanes}. However, since polygonal meshes are the most widely used 3D representation in downstream tasks, an additional post-processing step, commonly achieved by Marching Cubes \cite{10.1145/37401.37422}, is adopted to extract target mesh from the implicit field. This step often encounters issues such as visual artifacts, adversely affecting the quality of the final reconstructed mesh.

\subsection{Mesh Reconstruction Method}
How to reconstruct high-quality meshes from multi-view images has been a long-standing problem in Computer Vision. Classically, this is addressed by photogrammetry pipelines which integrate structure-from-motion (SfM) \cite{schonberger2016structure} and multi-view stereo (MVS) \cite{schonberger2016pixelwise} techniques. However, these traditional mesh reconstruction methods are tailored for dense-view and real-captured input images, making them unsuitable for sparse inconsistent images generated by current multi-view diffusion models. Therefore, many previous 3D generation works \cite{long2023wonder3d,liu2023syncdreamer, liu2023one2345, li2023instant3d} obtain the final polygonal mesh by exporting NeRF representation to mesh-based representation with surface extraction such as Marching Cubes (MC) \cite{10.1145/37401.37422}.
 
 Recently, with the advancement of machine learning, many works explore gradient-based mesh reconstruction schemes, which extract surface from an implicit function encoded via convolutional networks and evaluate objectives on the mesh. DMTet \cite{shen2021deep} utilizes a differentiable marching tetrahedra layer that converts the implicit signed distance function to explicit mesh. However, vertices of extracted mesh in DMTet are unable to move independently, leading to surface artifacts. FlexiCubes \cite{shen2023flexicubes} integrates flexibility to the mesh-based representation by introducing extra weight parameters. Such gradient-based surface extraction methods combined with differentiable rendering enable an end-to-end sparse-view mesh reconstruction training.

\begin{figure}[tb]
  \centering
  \vspace{-1cm}
  \includegraphics[width=1.0\textwidth]{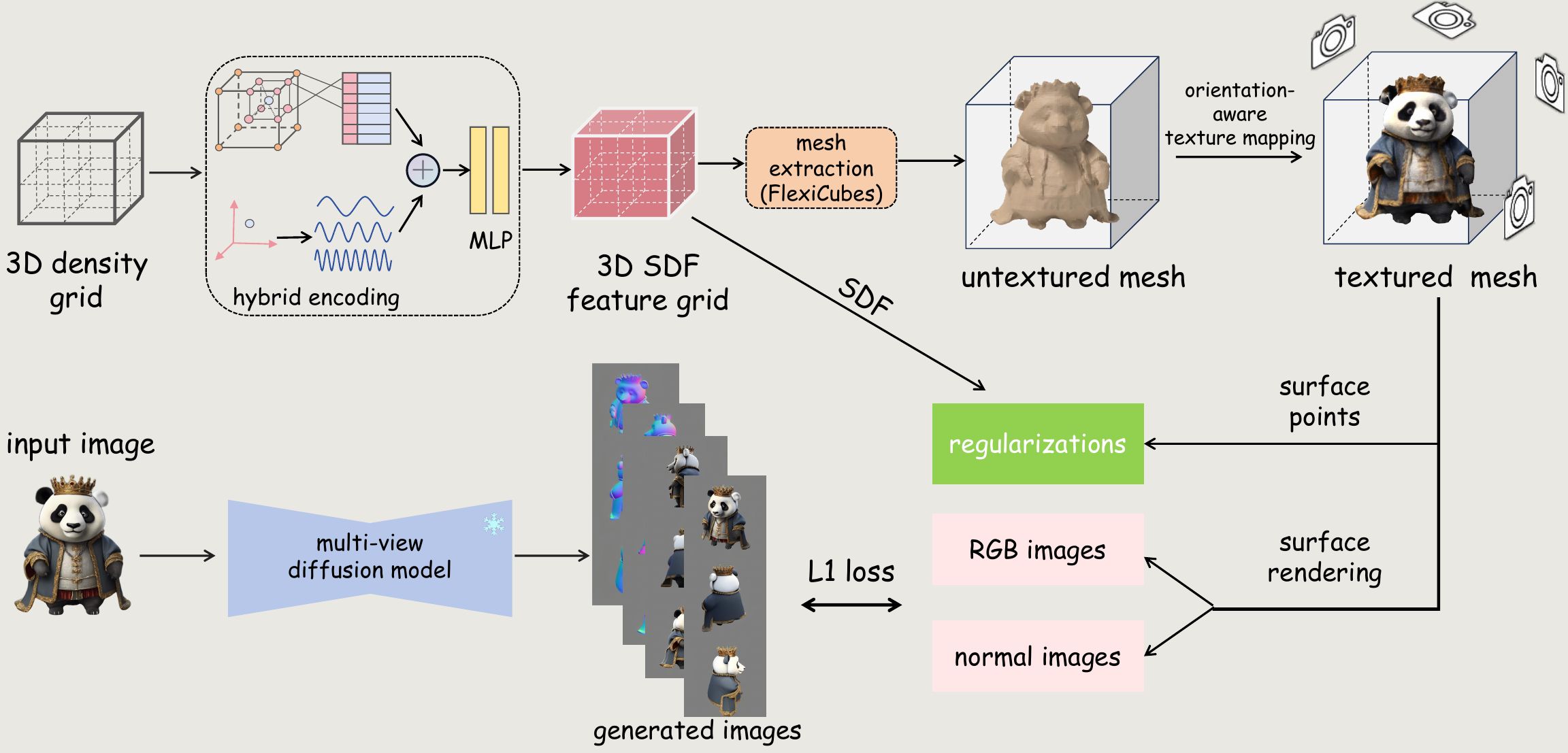}
  \centering
  \caption{\textbf{The pipeline of FlexiDreamer}. The inference image is fed into multi-view diffusion model to generate multi-view images. Then an end-to-end reconstruction framework based FlexiCubes is trained end-to-end for a high-quality mesh. The mesh extracted from signed distance function can be iteratively optimized by minimizing the difference between its rendering images and multi-view generated images.
  }
  \label{fig:pipeline}
\end{figure}

\section{Method}

In this section, we present a detailed overview of FlexiDreamer (illustrated in Figure \ref{fig:pipeline}). Given a single input image, our model first utilizes a pre-trained multi-view diffusion model (see Sec \ref{diffusion}) to generate multi-view images. Then we develop our end-to-end mesh reconstruction framework (see \ref{reconstruction}) to reconstruct 3D high-quality textured mesh from multi-view images. Finally, we describe the objective function to train our reconstruction framework (see Sec \ref{loss}).

\subsection{Multi-view Images Generation with Diffusion Model} \label{diffusion}


As illustrated in Figure \ref{fig:pipeline}, we adopt a two-step image-to-3D pipeline. We firstly take advantage of a single image to a pre-trained multi-view diffusion model to generate multi-view RGB images. Specifically, we adopt Zero123++ \cite{shi2023zero123}, which is a high-performance diffusion model for multi-view images generation, to produce 6 images with evenly distributed azimuths and interleaving elevations of $-10^\circ$ and $20^\circ$. Since pure RGB images supervision may bring geometry degradation (See \ref{normal} for details), we generate corresponding normal images from RGB images by a pre-trained ControlNet \cite{zhang2023adding} to provide additional geometry information supervision for mesh reconstruction.

\subsection{End-to-end Mesh Reconstruction Framework} \label{reconstruction}
 
\subsubsection{Architecture based on FlexiCubes}



Our mesh reconstruction framework is built upon FlexiCubes \cite{shen2023flexicubes}, which directly optimizes the textured mesh by minimizing the difference with multi-view generated images. During each training iteration, we employ a neural network to estimate the signed distance function (SDF) for a pre-defined 3D density grid space. Then we use FlexiCubes to extract an untextured mesh according to the zero-level set of the signed distance function by Dual Marching Cube algorithm \cite{nielson2004dual}. To obtain the texture of mesh, we query the colors of vertices from multi-view images by surface rendering \cite{Laine2020diffrast}. The textured mesh can be iteratively optimized with the neural signed distance function network parameters through backpropagation, driven by the losses between mesh's rendered images and multi-view generated images. With these designs, our mesh reconstruction framework supports an end-to-end training, yielding a high-quality textured mesh as the final output.

\subsubsection{Hybrid Positional Encoding in SDF Learning} 


To achieve computational efficiency during training, we adopt Instant-NGP \cite{mueller2022instant} to represent signed distance function (SDF). Instant-NGP can fasten the convergence of SDF geometric fitting by utilizing a multi-scale hash encoding to encode positions into learnable features of a hash table. In this encoding scheme, a given point $\boldsymbol{x}$ is mapped to its corresponding position at each grid resolution and its feature is obtained through tri-linear interpolation. Features from all resolutions are concatenated together to form $\boldsymbol{x}$'s hash grid encoding features $\gamma_h(\boldsymbol{x})$.

However, the geometry representation based on discrete grids in Instant-NGP often leads to non-smooth and distorted reconstruction surface with the supervision of sparse and inconsistent multi-view images. To enhance reconstruction quality, we provide more geometric priors by designing a hybrid positional encoding:

We combine fourier encoding features of $\boldsymbol{x}$ with its hierarchical hash grid features to mitigate the geometry distortion issue. Specifically, we firstly feed $\boldsymbol{x}$ into sinusoidal functions with low frequency levels ($m$ levels) to obtain its fourier features $\gamma_f(\boldsymbol{x})$:

\begin{equation}
    \gamma_f(\boldsymbol{x}) = [\mathrm{sin}(\boldsymbol{x}), \mathrm{cos}(\boldsymbol{x}), \cdots, \mathrm{sin}(2^{m-1}\boldsymbol{x}), \mathrm{cos}(2^{m-1}\boldsymbol{x})]
\end{equation}

Then we input the concatenation of $\gamma_h(\boldsymbol{x})$ and $\gamma_f(\boldsymbol{x})$ into a tiny MLP characterized by shallow layers and low-dimensional hidden units. Therefore, the computational load of our neural network with this combined positional encoding remains relatively low compared to Instant-NGP and our network is still able to realize a fast convergence. Furthermore, to balance the combination of two features, we multiply hash grid features by a ratio $\alpha$ (see Sec \ref{choice_of_alpha} for details):

\begin{equation}
    \gamma_{b}(\boldsymbol{x}) = (\gamma_f(\boldsymbol{x}), \alpha \cdot \gamma_h(\boldsymbol{x}))
\end{equation}

to obtain the final hybrid positional encoding features $\gamma_{b}(\boldsymbol{x})$. Moreover, the addition of low-level fourier encoding features also helps capture more surface details, because it improves the limited representation power of the coarse-level grids of the multi-resolution hash grid encoding (see Sec \ref{hybrid} for details).

\subsubsection{Orientation-aware Texture Mapping}
We also adopt an orientation-aware texture mapping strategy to tolerate inconsistencies across different views of generated RGB images. Specifically, we first choose the nearest viewpoint of 6 views for each surface point on mesh by measuring the angle between their vertex normal and viewing direction. The nearest viewpoint for each surface point is determined by the smallest angle between its normal and the viewing direction. Then we assign the pixel color in the nearest-view image to each point based on its texture coordinate. This configuration helps mitigate negative effects from the inaccuracies at the overlap area of two adjacent views on the textured mesh.

\subsubsection{Regularizations for Better Geometry} \label{regularization}

We further improve the reconstruction quality by incorporating some extra regularizations into FlexiCubes. Lacking in dense and rich-textured supervised multi-view images, we incorporate additional geometric priors to our mesh reconstruction framework by applying an eikonal regularization:

\begin{equation} 
\mathcal{R}_{eikonal} = \frac{1}{K}\sum_{i=1}^{K}(\left\| \nabla f\left(\boldsymbol{x}_i\right) \right\| - 1)^2
\end{equation}

where $K$ is the number of points in the 3D density grid while $f$ represents the signed distance function. This design is inspired from the volumetric rendering in NeuS \cite{wang2023neus}, where it is employed to constrain the magnitude of SDF gradients to be unit length. In this work, we verify its effectiveness in our surface rendering procedure of mesh (see Sec \ref{eikonal} for details). Moreover, since the surface of extracted mesh by FlexiCubes is rough, we adopt a laplacian regularization $\mathcal{R}_{laplacian}$ inspired by \cite{luan2021unified}. It helps make the surface smoother by minimizing each vertex's distance to the average position of its neighbours. We also integrate a normal consistency regularization $\mathcal{R}_{consistency}$ to further smooth the mesh by minimizing the negative cosine similarity between connected face normals (see Sec \ref{smooth} for details).



\subsection{Training Objectives} \label{loss}

The overall objective function of FlexiDreamer is defined as:

\begin{equation}
   \mathcal{L}_{total} = \mathcal{L}_{rgb} + \mathcal{L}_{mask} + \mathcal{L}_{normal} + \mathcal{R}_{eikonal} 
   + \mathcal{R}_{laplacian} + \mathcal{R}_{consistency} 
\end{equation}
where we compute a L1 loss between the rendered images and multi-view images for $\mathcal{L}_{mask}$, $\mathcal{L}_{normal}$ and $\mathcal{L}_{rgb}$. $\mathcal{R}_{eikonal}$, $\mathcal{R}_{laplacian}$ and $\mathcal{R}_{consistency}$ discussed in Sec \ref{regularization} are also utilized to avoid undesired reconstruction artifacts.

\section{Experiments}

\subsection{Implementation Details} \label{details}

In practice, the pre-trained multi-view diffusion model in FlexiDreamer is adopted from Zero123++\cite{shi2023zero123}, because it has a high performance for multi-view image generation task. 

We train our reconstruction framework for 600 iterations and utilize an Adam optimizer to optimize the neural networks for signed distance function and texture with learning rates of 1e-3 and 1e-2 respectively. The network to represent signed distance function has three-layers of MLP while the network to represent texture has one layer. The geometry constraints weights, denoted as $\lambda_{normal}$ and $\lambda_{mask}$, are configured to 2.0 and 15.0 respectively for geometry supervision, and $\lambda_{rgb}$ is carefully set to 0.1 to supervise a more realistic texture. The weight assigned to the eikonal regularization, denoted as $\lambda_{eikonal}$, is set to 0.1. Additionally, we set $\lambda_{laplacian}$ to 0.4 and $\lambda_{consistency}$ to 0.05 for laplacian regularization and normal consistency regularization to obtain a smoother mesh surface. The grid resolution of 3D density grid is set to 96 to strike a balance between generation speed and quality (see Sec \ref{resolution} for details).

\begin{figure}[tb]
  \centering
  \includegraphics[width=1.0\textwidth]{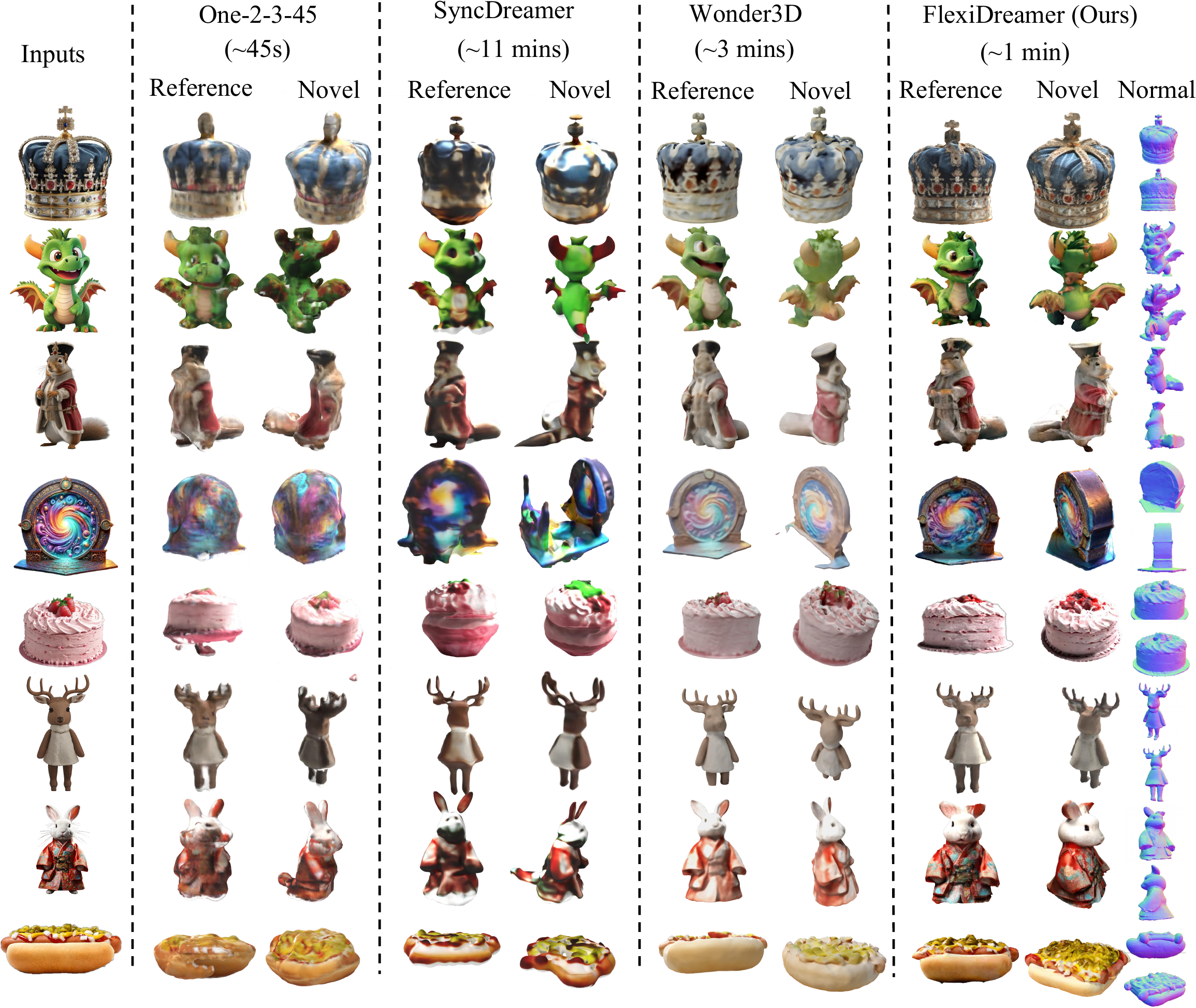}
  \centering
  \caption{
   The qualitative comparisons with baselines in terms of the generated textured meshes. It reveals a superior performance of FlexiDreamer in reconstructing both geometry and texture details from single-view images. 
  }
  \label{fig:compare}
\end{figure}

\subsection{Qualitative Results} \label{qualitative}

To illustrate FlexiDreamer's scalability on single image-to-3d task, we qualitatively compare its performance with previous works. The implementation is based on their official open-source code with default parameters. 

Figure \ref{fig:compare} visualizes some examples of the shapes generated from single images in geometry, texture and speed. It can be seen that compared to baselines, our method produces sharper geometric details, more distinct textures, and more consistent surfaces in a short amount of time. This is because our mesh reconstruction model based on FlexiCubes \cite{shen2023flexicubes} optimizes the mesh by fast surface rendering and integrates more priors to reconstruct geometry from multi-view generated images. Additionally, our method also enables an end-to-end training with textured mesh as the final output, thus bypassing slow post-extraction adopted in baselines.

In Appendix, we show more comparisons with baselines and more results of high-fidelity textured meshes generated by our method.

\begin{table}[tb]
  \caption{Quantitative comparisons for geometry and texture quality between our method and baselines for single image-to-3D task on Google Scanned Object (GSO) \cite{downs2022google} dataset. 
  }
  \label{tab:iou}
  \centering
  \tabcolsep=0.1cm
  \begin{tabular}{@{}lllllll@{}}
    \toprule
    Method & Chamfer Dist$\downarrow$ & Volume IoU$\uparrow$ & F-score ($\%$) $\uparrow$ &  LPIPS$\downarrow$ & Clip-Sim $\uparrow$ & Time  \\
    \midrule
    Wonder3D\cite{long2023wonder3d}  & 0.0186 & 0.4398 & 76.75 &  0.2554 &  83.70 & 3 min\\
    SyncDreamer\cite{liu2023syncdreamer} & 0.0140 & 0.3900 & 75.74  & 0.2591 & 82.76 & 11 min\\
    One-2-3-45\cite{liu2023one2345} & 0.0172 & 0.4463 & 72.19 &  0.2625 & 79.83 & 45 s\\
    Magic123\cite{qian2023magic123} & 0.0188 & 0.3714 & 60.66 &  0.2442 & 85.16 & 1 h\\
    LGM \cite{tang2024lgm} & 0.0117 & 0.4685 & 68.69 &  0.2560 & 85.20 & 1 min\\
    Ours &  \textbf{0.0098} & \textbf{0.5078} & \textbf{78.23} & \textbf{0.2398} & \textbf{87.63} & 1 min\\
  \bottomrule
  \end{tabular}
\end{table}

\subsection{Quantitative Results} \label{quantitative}

We choose Google Scanned Object (GSO) dataset \cite{downs2022google} to evaluate our method. Firstly, we scale all the generated and ground-truth meshes to an identical size and align their centers at the same position. We also carefully adjust their poses for more accurate evaluation. Subsequently, we employ Chamfer Distances (CD), Volume IoU and F-score (with a threhold of 0.05) to assess the similarity in geometry. Results on the left side of table \ref{tab:iou} demonstrates that our method outperforms all the baselines in reconstructing accurate geometric shapes.

Additionally, to evaluate the quality of mesh texture, we respectively render 24 images from ground-truth and generated meshes. Then we use LPIPS\cite{zhang2018unreasonable} and Clip-Similarity metrics to measure the resemblance of their appearance. Results on the right side of table \ref{tab:iou} summarizes that our method outperforms all the baselines in recovering realistic texture.

\subsection{Abalation Study}

\begin{figure}[htb]
  \centering
  \vspace{-1cm}
  \includegraphics[width=1.0\textwidth]{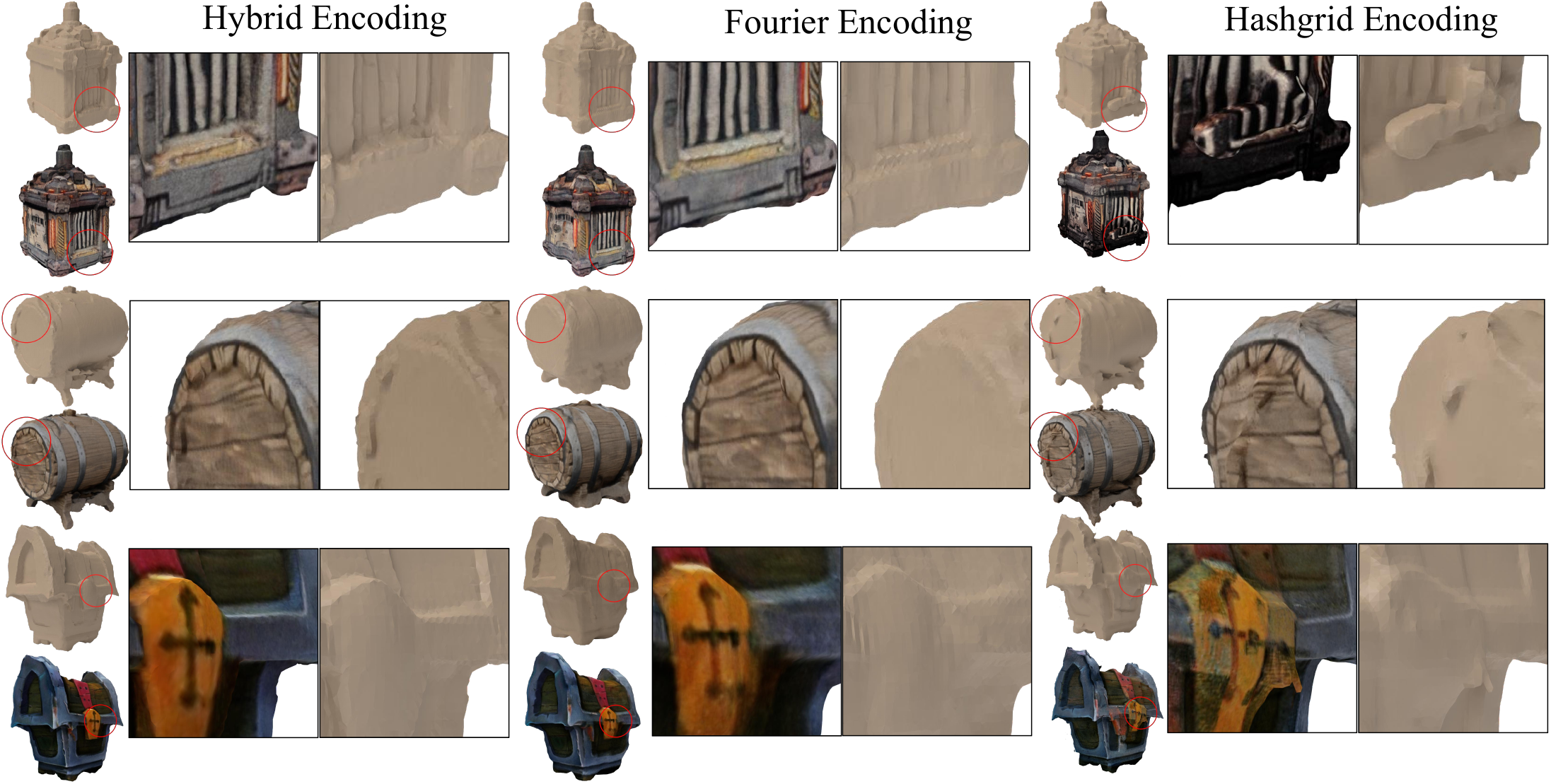}
  \centering
  \caption{Results of using different encoding scheme. Compared to fourier encoding and hashgrid encoding scheme, our model with hybrid encoding generates the most accurate geometry and texture.}
  \label{fig:hybrid}
\end{figure}

\subsubsection{Design of Hybrid Positional Encoding} \label{hybrid}

We examine the feasibility of adopting a hybrid position encoding scheme in signed distance function (SDF) learning to reduce geometry distortions. We respectively integrate fourier positional encoding, hybrid positional encoding and  multi-resolution hash grid encoding into the SDF neural network and compare the geometry of their final mesh results. In Figure \ref{fig:hybrid}, we can find that both fourier encoding and hash grid encoding are unable to reconstruct an accurate geometry from multi-view images, because the former generates over-smooth surface (more clearly in the second example) and the latter makes reconstruction shapes distorted. In contrast, the hybrid encoding can mitigate the geometry distortion and generate a more detailed surface. 

\subsubsection{Orientation-aware Texture Mapping Strategy}

\begin{figure}[tb] 
  \centering
  \includegraphics[width=1.0\textwidth]{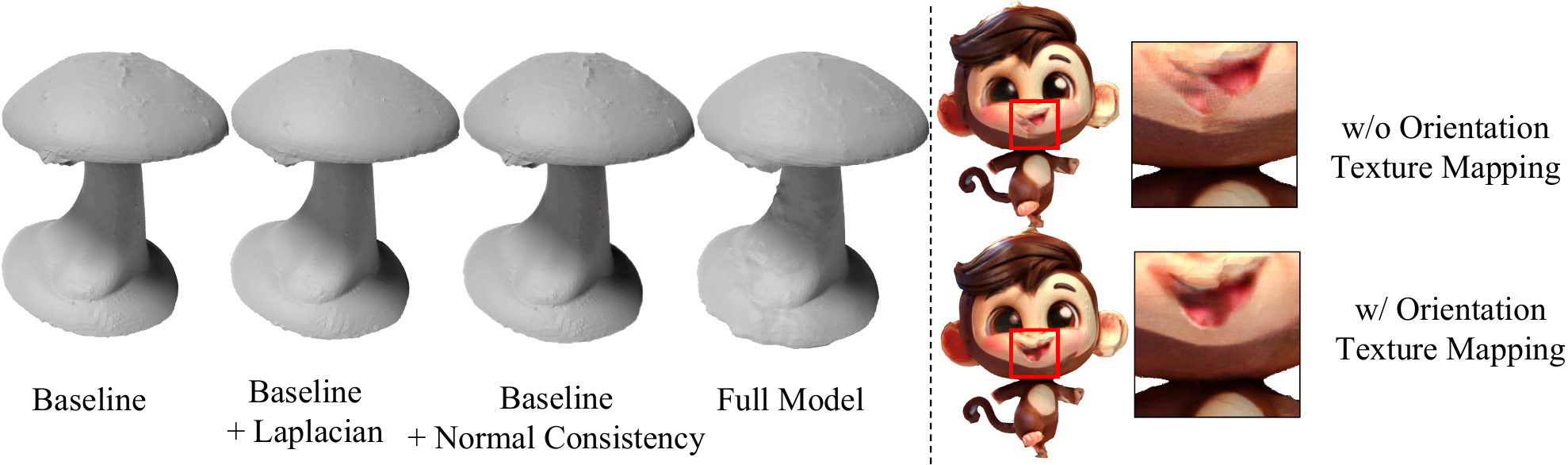}
  \caption{
   Ablation study on the smooth regularizations and orientation texture mapping strategy. For smooth regularization, the laplacian smoothing term can smooth the surface of the mesh at a global scale while normal consistency constraint helps reduce high-frequency noises. For orientation texture mapping, it can be seen that it is helpful for mitigating surface ghosting.
  }
  \label{fig:smooth_ori}
\end{figure}

We validate the effectiveness of our proposed orientation-aware texture mapping. To compare, we train our reconstruction framework with a trivial texture mapping. The results are illustrated in the right side of Figure \ref{fig:smooth_ori}. It can be figured out that the meshes without using orientation-aware texture mapping have severe ghosting problems on the surface, due to the inconsistent details existing in supervised RGB images. On the contrary, the meshes reconstructed with orientation-aware texture mapping strategy have a more accurate appearance. However, this strategy is unable to fully address the inconsistencies of multi-view images. We will explore to further overcome this challenge for future work.

\subsubsection{Smooth Regularization} \label{smooth}

To assess the effectiveness of the two smooth regularizations we employed, which are laplacian regularization and normal consistency regularization, we conduct experiments in generating a mushroom featuring a smooth cap. The visualization results are depicted in the left side of Figure \ref{fig:smooth_ori} (please zoom in to see them more clearly). It is noticed that the baseline model's surface exhibits severe unevenness. Incorporating either laplacian smoothness or normal consistency constraints can help mitigate the noisy surfaces. Specifically, the laplacian penalty term is adept at smoothing the mesh surface at a global scale. Meanwhile, the normal consistency term plays a pivotal role in diminishing high-frequency noises, thereby enhancing the overall quality of the mesh structure. Finally, the combination of both smoothing strategies yields optimal performance, resulting in clean surfaces while retaining intricate details.

\subsubsection{Eikonal Regularization} \label{eikonal}

To assess the effectiveness of eikonal regularization integrated in the model, we respectively train FlexiDreamer with and without $\mathcal{R}_{eikonal}$. The comparison of reconstruction results are shown in Figure \ref{fig:eikonal_normal} (a). We can see that the meshes generated without eikonal constraints have geometric holes or outliers. This illustrates that eikonal regularization is helpful for our model, which is featured by surface rendering, in constraining signed distance function to represent a complete geometry.

\subsubsection{The Importance of Normal Loss} \label{normal}

\begin{figure}[tb] 
  \centering
  \includegraphics[width=1.0\textwidth]{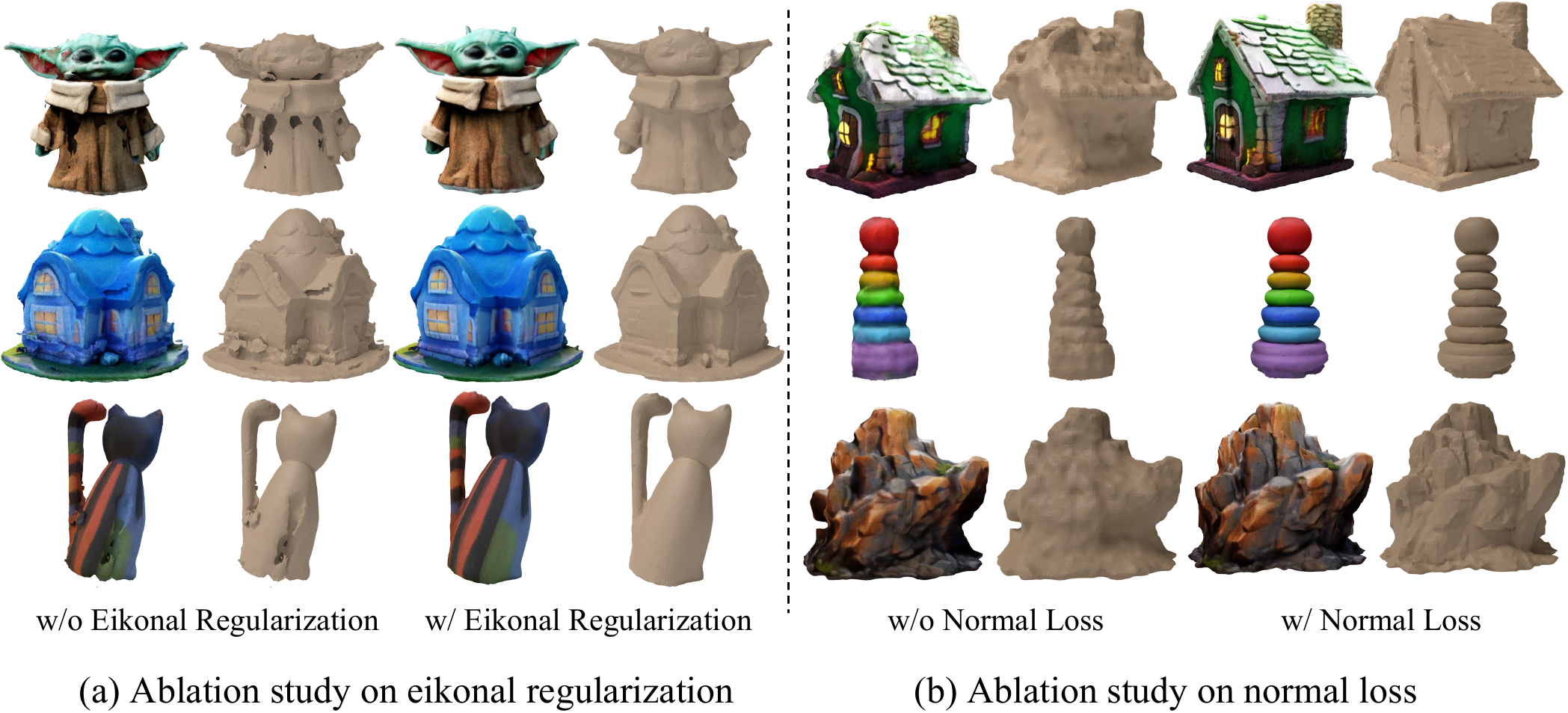}
  \caption{
   Ablation study on eikonal regularization and normal loss respectively. (a) Without employing eikonal regularization, the reconstruction geometry has holes or outliers. (b) Without supervision of normal images, the geometric shapes of results are not correct, and sometimes degrade.
  }
  \label{fig:eikonal_normal}
\end{figure}

We examine the significance of normal images supervision to reconstruction results. To compare, we train FlexiDreamer without  $\mathcal{L}_{normal}$. The results are shown in Figure \ref{fig:eikonal_normal} (b). It is evident that the results generated without normal constraints suffer from severe geometry degradation. This is because normal images can provide crucial geometric information for the reconstruction process, especially when dealing with complex geometric shapes.

\section{Conclusion}

In this paper, we introduce FlexiDreamer, a novel framework for generating high-quality textured meshes from single-view images. By utilizing a gradient-based mesh optimization method FlexiCubes, FlexiDreamer realizes a rapid end-to-end acquisition of target textured meshes without extra procedure like Marching Cubes. We adopt a hybrid encoding scheme and a texture mapping strategy to tolerate the inconsistencies from multi-view generated images respectively in reconstruction geometry and texture. To further enhance the reconstruction results, we integrate extra regularizations into our mesh reconstruction framework for better geometry. Overall, FlexiDreamer can generate realistic textured meshes in approximately 1 minute.


\subsection{Limitations} \label{limitations}
Given that our model is essentially a multi-view reconstruction model, the quality of 3D generated meshes is heavily dependent on the quality of multi-view images. Due to the limited ability of current multi-view diffusion model to generate dense and consistent images, some reconstruction results of FlexiDreamer are not satisfactory. Additionally, our model also struggles to generate accurate 3D geometric shapes from input images with large elevation angles. Finally, The resolution of our density grid can only be set up to 128 due to the limited computing resources, which hinders its ability to generate more detailed surface.

\subsection{Potential Negative Impact} \label{negative}
Although our method excels in producing hyper-realistic 3D assets within a short amount of time, the high fidelity of our generated results might raise a concern about misuse by malicious entities. Therefore, it is necessary for developers and users to remain vigilant to prevent the occurrence of such harmful practices.

\bibliographystyle{plain}
\bibliography{neurips_2024}

\appendix
\newpage

\section{More Implementation Details}

For multi-view image generation, we adopt Zero123++ \cite{shi2023zero123} as our multi-view diffusion model. The guidance of Zero123++ is configured to 4 while the number of inference steps is set to 75. We also utilize the ControlNet integrated in Zero123++ to generate normal images from its multi-view RGB outputs.

In our hybrid encoding scheme, the level of fourier encoding $m$ is 6 and the resolution of hierarchical hash grids is chosen to be a geometric progression between the coarsest and finest resolutions $[2^4, 2^{10}]$ with 12 levels. Each hash entry has a channel of 2. The combination ratio of hash grid features $\alpha$ equals to 0.1.

Our implementation is built on the official code of Zero123++ and FlexiCubes\cite{shen2023flexicubes}. For Sec \ref{qualitative} and \ref{quantitative}, we compare FlexiDreamer with several competitive baselines by using their official code on GitHub, including One-2-3-45\cite{liu2023one2345}, SyncDreamer\cite{liu2023syncdreamer}, Wonder3D\cite{long2023wonder3d} Magic123\cite{qian2023magic123} and LGM\cite{tang2024lgm}.

\section{Camera Settings}

We inherit the camera settings from Zero123++: In order to circumvent the error in the elevation estimation module incorporated in previous pipelines, such as One-2-3-45 \cite{liu2023one2345} and DreamGaussian \cite{tang2023dreamgaussian}, we use camera poses with relative azimuth and absolute elevation angles to the input view for novel view synthesis to minimize the orientation inconsistency. To be more specific, the six poses comprise interleaving elevations of $-10^\circ$ and $20^\circ$, accompanied by azimuths that span from $30^\circ$ to $330^\circ$ with an increase by $60^\circ$ for each pose. These carefully chosen camera poses can totally scan the whole object. Moreover, we adopt a perspective camera system and the camera is positioned at a distance of 3.0 meters from the coordinate origin, i.e. the radial distance is 3.0.

\section{Additional Ablation Studies}

\begin{figure}[htb]
 \centering
 \hspace{-8mm}
 \begin{minipage}[c]{0.4\textwidth}
  \centering
  \includegraphics[width=1.1\textwidth]{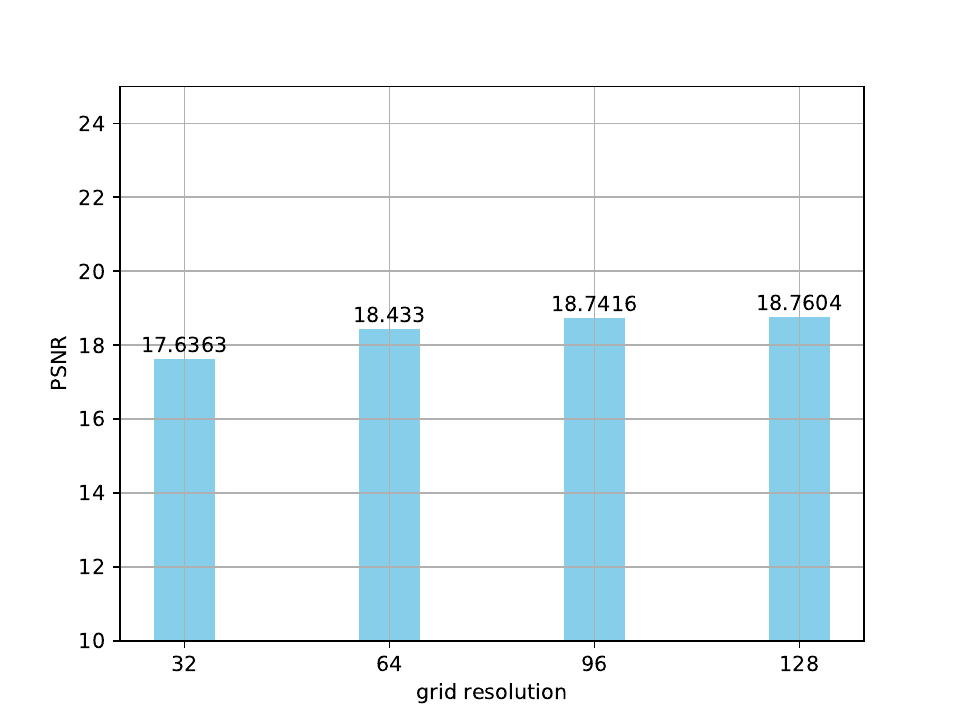}
 \end{minipage}
 \hspace{15mm}
 \begin{minipage}[c]{0.5\textwidth}
  \centering
  \includegraphics[width=0.9\textwidth]{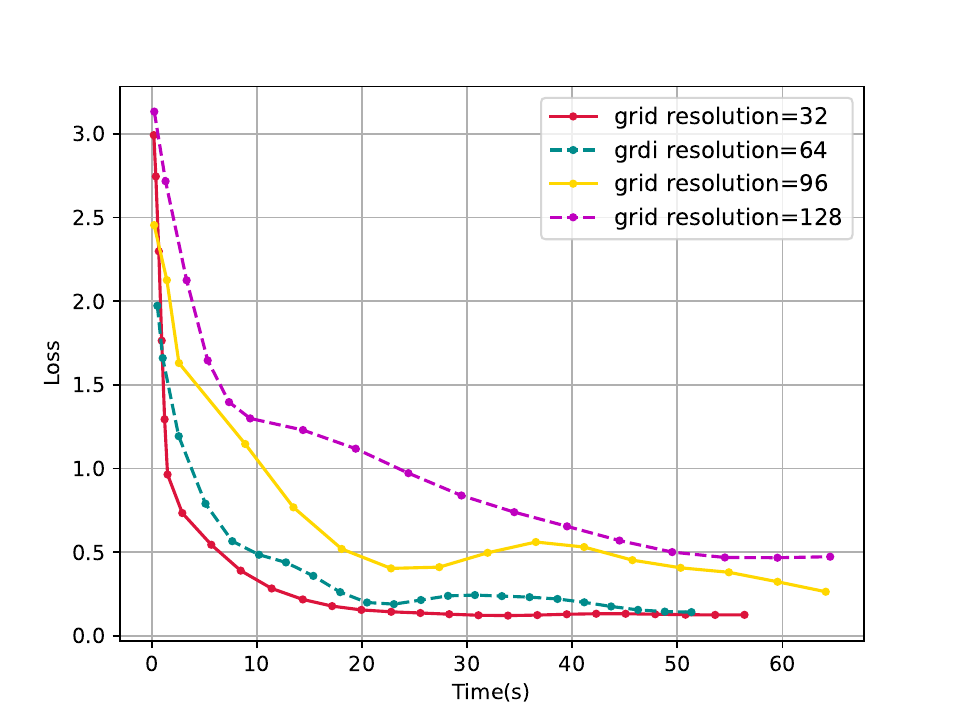}
 \end{minipage}
 \caption{
   Convergence curves of the model with different training grid resolutions. PSNR metrics of the generated meshes surface are also compared. 
   }
    \label{fig:add_ablation2}
\end{figure}

\subsection{Training Grid Resolution} \label{resolution}

The resolution of the 3D density grid depicted in Figure \ref{fig:pipeline} determines the computational complexity and details of mesh surface. In order to choose a suitable resolution for the grid, we conduct experiments using various resolutions, namely 32, 64, 96, and 128. As illustrated in Figure \ref{fig:add_ablation2}, we plot curves of loss under different settings throughout the training process. We also evaluate PSNR metrics of the images rendered from the final results. It is can be seen that although the quality of reconstruction results enhances with increasing resolution, our model exhibits a slower convergence. Moreover, the convergence time at the resolution of 128 is significantly longer than the other three resolutions. However, compared with resolution of 96, where the model already has an acceptable performance, the improvements achieved at resolution of 128 are marginal. Therefore, we select 96 as our grid resolution is selected to strike a balance between generation speed and quality.

\begin{figure}[tb]
  \centering
    \vspace{-1.5cm}
  \includegraphics[width=1.0\textwidth]{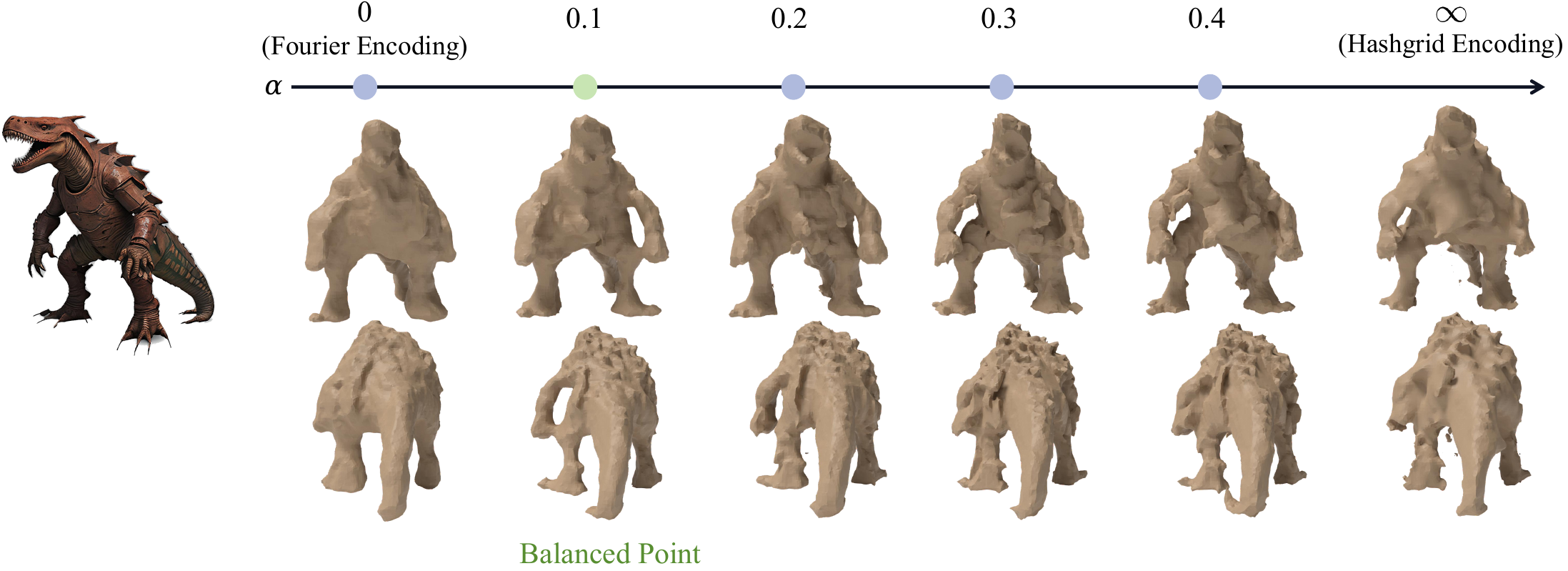}
  \centering
  \caption{
  We study the effects of $\alpha$ on our model. Increasing $\alpha$ leads to a 3D geometry with more surface details but less precision. We find $\alpha = 0.1$ provides a good balance.
  }
  \label{fig:alpha}
\end{figure}

\subsection{Choice of Combination Ratio $\alpha$} \label{choice_of_alpha}

 We conduct experiments with an increasing value of $\alpha$ and compare their reconstruction geometry. We start from $\alpha = 0$ to use only fourier encoding and gradually increase $\alpha$ to 0.1, 0.2, 0.3, 0.4, and finally $\infty$ to use only multi-resolution hash grid encoding. The comparison of results is illustrated in Figure \ref{fig:alpha}.  The key observations include: (1) Relying solely on fourier encoding generates over-smooth mesh surface  with minimal details. (2) Relying solely on the hash grid encoding improves performance in recovering details, but generates a distorted shape. (3) As $\alpha$ increases, the reconstructed geometry has more surface details but becomes more inaccurate. To strike a balance, we choose 0.1 for $\alpha$.

\begin{figure}[hb]
  \centering
  \includegraphics[width=1.0\textwidth]{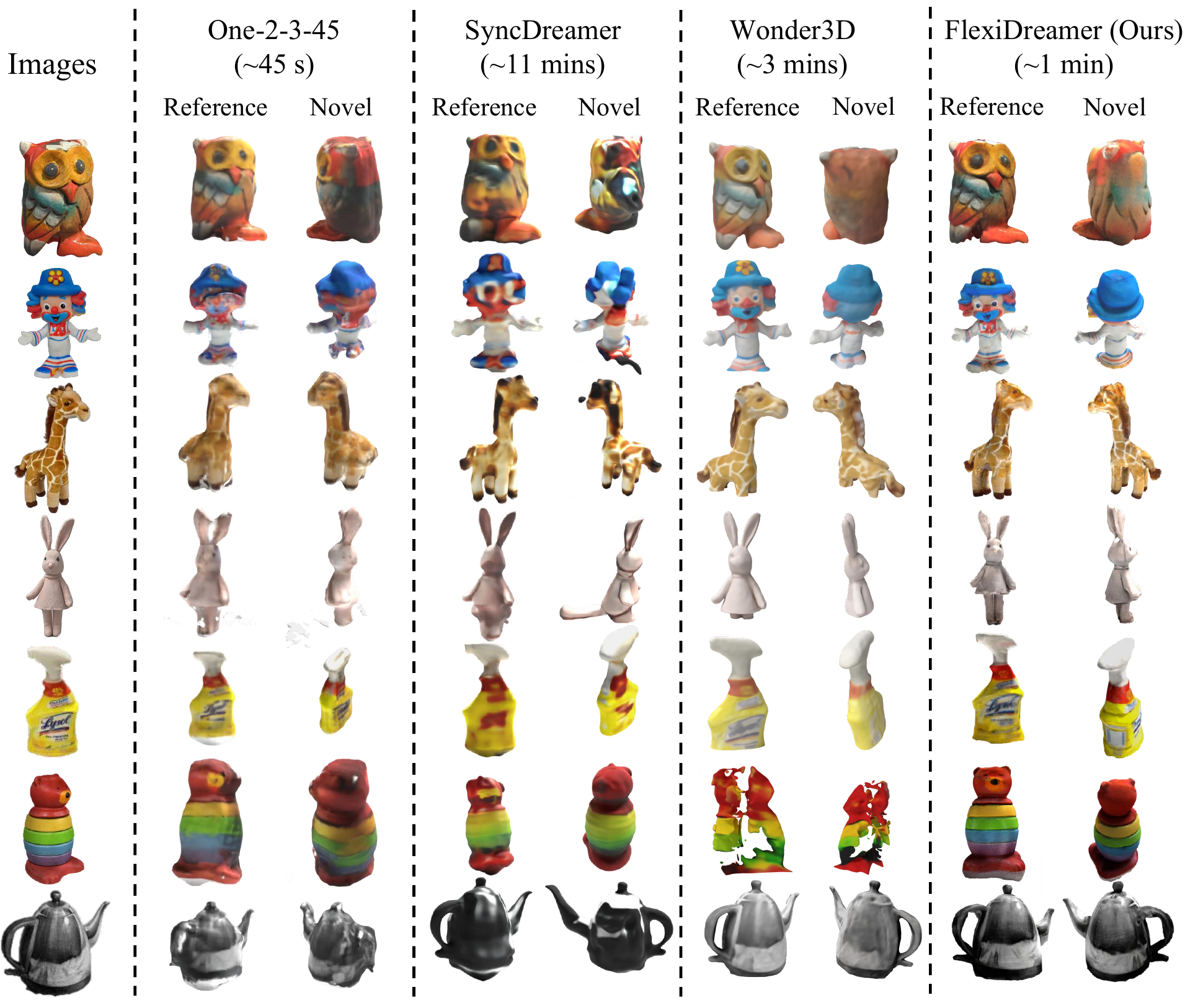}
  \centering
  \caption{
  \textbf{Qualitative comparisons:} We compare FlexiDreamer with recent state-of-art single image-to-3D approaches.
  }
  \label{fig:add_comparison}
\end{figure}

\section{More Comparisons and More Results}
\begin{figure}[hb]
  \centering
    \vspace{1cm}
  \includegraphics[width=1.0\textwidth]{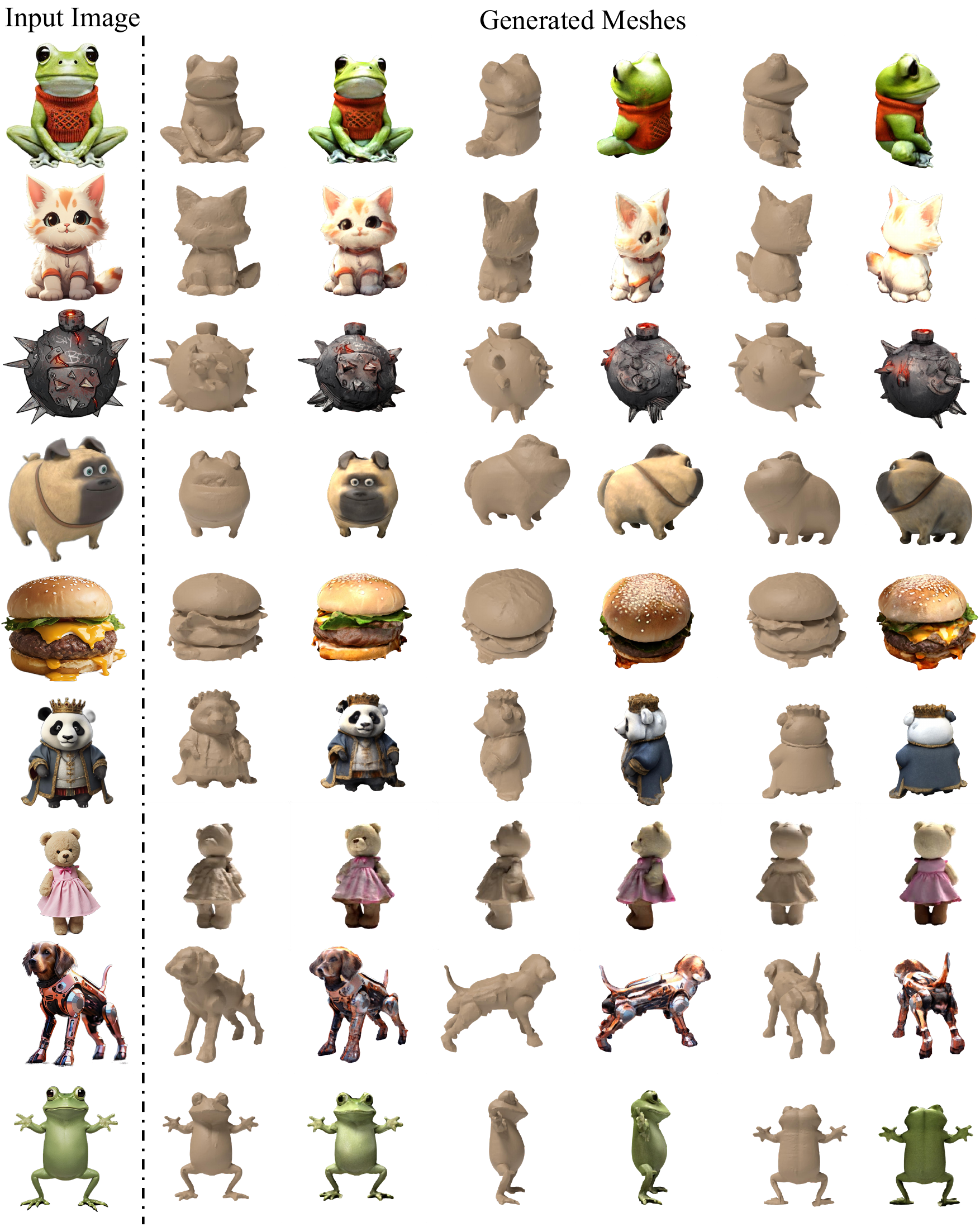}
  \centering
  \caption{
  More generated results of FlexiDreamer.
  }
  \label{fig:more_results}
\end{figure}

We additionally compare FlexiDreamer with baselines and show more visualization results respectively in Figure \ref{fig:add_comparison} and \ref{fig:more_results}.

\end{document}